\documentclass[runningheads]{llncs}

\usepackage{eccv}

\usepackage{eccvabbrv}

\usepackage{graphicx}
\usepackage{booktabs}
\usepackage{indentfirst}
\usepackage{amsmath}
\usepackage{multirow}
\usepackage{colortbl}
\usepackage{array}
\usepackage{makecell}
\usepackage{xcolor}
\usepackage{listings}

\definecolor{codegreen}{rgb}{0,0.6,0}
\definecolor{codegray}{rgb}{0.5,0.5,0.5}
\definecolor{codepurple}{rgb}{0.58,0,0.82}
\definecolor{backcolour}{rgb}{0.97,0.97,0.97}
\definecolor{keywordcolor}{rgb}{0.0,0.44,0.87}
\lstdefinestyle{academicsnippet}{
	backgroundcolor=\color{backcolour},
	commentstyle=\color{codegreen}\itshape,
	keywordstyle=\color{keywordcolor}\bfseries,
	numberstyle=\tiny\color{codegray}\sffamily,
	stringstyle=\color{codepurple},
	basicstyle=\ttfamily\scriptsize,
	breakatwhitespace=false,
	breaklines=true,
	captionpos=b,
	keepspaces=true,
	numbers=left,
	numbersep=8pt,
	showspaces=false,
	showstringspaces=false,
	showtabs=false,
	tabsize=4,
	frame=tb,
	rulecolor=\color{black!70},
	language=Python
}

\usepackage[accsupp]{axessibility}  

\usepackage{hyperref}

\usepackage{orcidlink}

\begin{document}

\title{Again-Pose: Anchor-Guided Adaptive Inter-Frame Motion Cues Propagating for High-quality Human Pose Reconstruction} 

\titlerunning{Again-Pose}

\author{Shuaikang Zhu\inst{1,2} \and
Yiding Sun\inst{2} \and
Yang Yang\inst{1}\thanks{Corresponding author.}}

\authorrunning{S.~Zhu et al.}

\institute{School of Electronic and Information Engineering, Xi'an Jiaotong University\\
\email{yyang@mail.xjtu.edu.cn}
\and
School of Software Engineering, Xi'an Jiaotong University\\
\email{zsk2001cn@stu.xjtu.edu.cn}}

\maketitle
\begin{abstract}
Reconstructing continuous 3D human poses from unconstrained videos is challenging, especially in extreme motion scenarios involving severe motion blur and occlusion. 
Current state-of-the-art methods typically rely on implicit temporal attention to aggregate features across frames. However, under severe visual degradation, input features often suffer from collapse, rendering them indistinguishable from noise. In such cases, implicit aggregation fails to distinguish valid signals, leading to catastrophic reconstruction errors. To address this robustness gap, we propose a simple yet effective framework called {\textbf A}nchor-{\textbf g}uided {\textbf a}daptive {\textbf i}{\textbf n}ter-frame motion cues propagating (\textbf{Again-Pose}), reformulating pose estimation in degraded frames as a motion-guided recovery task. Instead of blindly smoothing features, we explicitly identify high-quality Anchor Frames based on feature saliency and propagate reliable kinematic cues to "inpaint" the poses of degraded intermediate frames. Specifically, a Dual-path Motion-aware Module captures fine-grained inter-frame dynamics, while a Difference-weighted Fusion Module adaptively propagates these cues to suppress drift. Extensive experiments on standard benchmarks (Human3.6M, 3DPW, PoseTrack) and the challenging FineDiving dataset demonstrate that Again-Pose significantly outperforms state-of-the-art methods in robustness and stability, effectively recovering plausible poses where other methods fail.
\end{abstract}
\section{Introduction}
\begin{figure}
	\centering
	\includegraphics[width=0.7\linewidth]{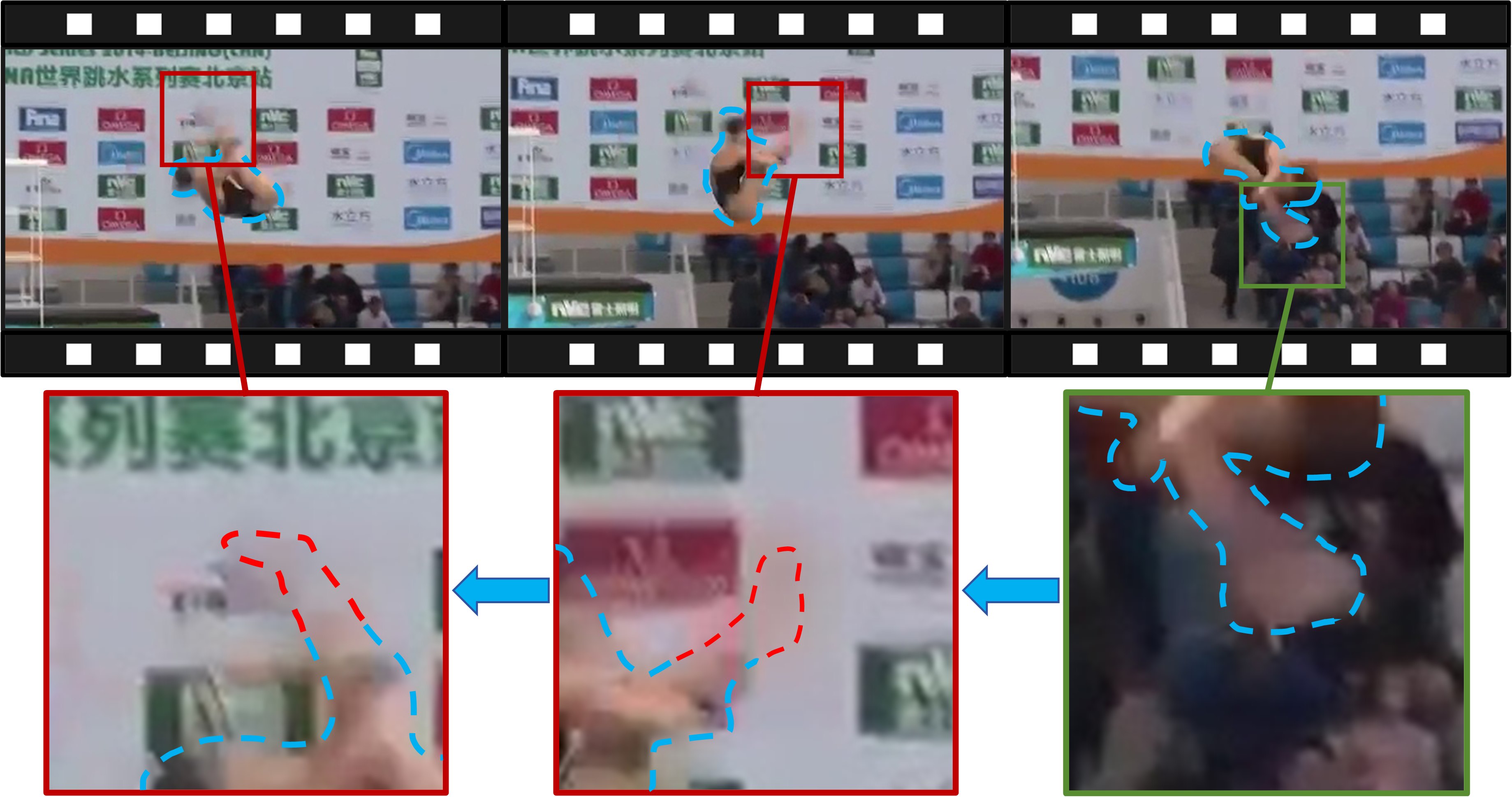}
	\caption{
		\textbf{Solving Feature Collapse via Anchor Propagation.} 
		In high-speed scenarios (\eg, diving), severe motion blur renders features in certain frames unreliable (\eg, the first two frames), leading to feature collapse. Our key insight is to treat pose estimation as a recovery task: we explicitly leverage high-quality Anchor Frames (\eg, the clear third frame) as reliable references. The clear body contour (\textbf{\textcolor{blue}{blue dashed line}}) from the anchor is propagated backwards via inter-frame motion dynamics to accurately infer the underlying shapes (\textbf{\textcolor{red}{red dashed line}}) in the preceding blurred target frames, effectively recovering visual evidence lost to blur.
	}
	\label{fig:Inspire}
\end{figure}
\noindent Reconstructing continuous 3D human poses from unconstrained video sequences is a fundamental problem in computer vision \cite{sun2025hyperpoint,wang2026pointrft,guo2026mantis}, holding immense value for applications such as augmented/virtual reality \cite{art-32}, human–robot interaction \cite{art-28, art-29, art-30, art-31}, and action quality assessment \cite{art-21, art-33}. Thanks to the development of parametric human body models (\eg, SMPL), significant progress has been made in pose estimation accuracy \cite{art-2, art-3, art-9}. Some methods achieve a Mean Per Joint Position Error (MPJPE) close to 50 mm on standard benchmarks. However, in real-world scenarios—especially extreme motion settings such as sports analytics \cite{art-34, art-35} and diving analysis \cite{art-21}—the primary challenge shifts from per-frame accuracy to robustness against degradation. In these scenarios, video frames are frequently plagued by severe \textbf{motion blur}, \textbf{abrupt pose changes}, and \textbf{occlusions}, causing standard reconstruction methods to fail catastrophically.\\
\indent Current human pose sequence reconstruction methods can be broadly divided into two categories: single-frame regression and temporal modeling. Single-frame-based methods (\eg, HMR 2.0 \cite{art-3}, HSMR \cite{art-2}) leverage large-scale ViT backbones to extract global features, achieving leading accuracy on benchmarks like PoseTrack \cite{art-17}. However, due to the lack of temporal constraints, they suffer from jitter and inconsistency when handling ambiguous frames. To address this, video-based methods introduce temporal modeling to enhance coherence. Approaches like VIBE \cite{art-27} employ recurrent architectures (GRU), while recent state-of-the-art methods like GLoT \cite{art-20}, and more recent works \cite{tram-2024, bio-2025}, utilize Transformers to capture long-range dependencies. These methods implicitly aggregate features across frames via attention mechanisms, significantly improving performance on standard datasets.\\
\indent However, a critical limitation remains overlooked: these temporal methods operate on the assumption that the input visual features are relatively reliable. They rely on implicit attention mechanisms to weigh temporal information. In extreme scenarios (\eg, high-speed diving or gymnastics), rapid motion causes severe \textbf{feature collapse} in blurry frames, where visual cues become indistinguishable. Under such conditions, implicit attention mechanisms struggle to distinguish between valid signals and noise, often propagating errors from low-quality frames throughout the sequence. As a result, even sophisticated Transformer-based models risk generating collapsed or chaotic meshes when the majority of a local window is degraded. To tackle this \textbf{robustness gap}, we argue that temporal information should not just be aggregated, but must be \textbf{explicitly filtered and propagated}. Instead of trusting all frames, we should identify "Anchors"—frames with high visual fidelity—and use them to recover the degraded frames. As illustrated in \cref{fig:Inspire}, although a single-frame might be severely blurred, the human body contour can be inferred from the inter-frame motion dynamics derived from adjacent high-quality anchors. This effectively turns the pose estimation problem in extreme scenarios into a motion-guided inpainting task. In this paper, we propose Again-Pose, an anchor-guided video pose estimation framework designed for robustness in extreme motion scenarios. The core idea is to decouple the sequence into reliable Anchor Frames and unreliable intermediate frames, balancing frame-wise accuracy with temporal continuity.\\
\indent Specifically, Again-Pose operates through three key components:
First, we design an \textbf{Intelligent Anchor-frame Selector}. Unlike heuristic keyframe selection, this mechanism computes a quality score combining visual saliency (sharpness) with a learnable confidence metric, explicitly filtering out collapsed features to establish a reliable foundation.
Second, we construct a \textbf{Dual-path Motion-aware Module}. It models inter-frame dynamics from two perspectives: parametric differences (pose changes) and visual dynamics (appearance flow), ensuring that the propagation respects the kinematic trends of the human body.
Finally, a \textbf{Difference-weighted Fusion Module} adaptively fuses predictions propagated from adjacent anchors based on temporal distance, suppressing drift and ensuring smooth transitions even across extended blurry intervals.
Extensive experiments on multiple benchmarks demonstrate that Again-Pose achieves state-of-the-art performance. Crucially, in the challenging FineDiving dataset characterized by high-speed motion, our method exhibits superior robustness compared to Transformer-based baselines, effectively recovering plausible poses where other methods fail. For instance, on the 3DPW dataset, our method improves the MPJPE metric by nearly 10 mm compared with the previous best-performing approach.\\
\indent The main contributions of this work are summarized as follows:\\
\indent (1) We identify the vulnerability of implicit temporal aggregation in extreme motion scenarios and propose Again-Pose, an explicit anchor-guided framework. To form a reliable foundation, we introduce an Intelligent Anchor-frame Selector that explicitly evaluates feature saliency to identify high-quality anchors and filter out degraded features.\\
\indent (2) We design a Dual-path Motion-aware Module coupled with a Difference-weighted Fusion strategy. It captures fine-grained inter-frame dynamics to explicitly propagate reliable kinematic cues from the identified anchors, effectively "inpainting" the poses of degraded intermediate frames.\\
\indent (3) Extensive experiments demonstrate that Again-Pose significantly outperforms state-of-the-art methods under severe motion blur and occlusion, providing superior temporal stability for downstream tasks like action quality assessment.

\section{Related Work}
\noindent\textbf{Single-frame-based Human Reconstruction.}\hspace{1em}
The task of single-frame human pose reconstruction primarily focuses on estimating SMPL model parameters directly from individual images. Early approaches like HMR \cite{art-4} pioneered the use of CNNs with adversarial priors to regress SMPL parameters. Subsequent works have improved upon this by incorporating iterative optimization (SPIN \cite{art-8}), mesh alignment feedback (PyMAF \cite{art-36, art-37}), and part-guided attention for occlusions (PARE \cite{art-22}). More recently, the introduction of Vision Transformers (ViT) has significantly boosted performance. HMR 2.0 \cite{art-3} leverages a large-scale ViT backbone to extract robust global features, setting a new standard for single-frame accuracy. HSMR \cite{art-2} further refines this by integrating biomechanical constraints. Non-parametric methods, such as GraphCMR \cite{art-39}, METRO \cite{art-40}, and Mesh Graphormer \cite{art-42}, directly regress mesh vertices. While they offer flexibility, they often lack the kinematic coherence provided by parametric models (SMPL) and struggle to maintain consistency in dynamic sequences without temporal constraints. In this work, we build upon the strong single-frame baseline of HMR 2.0 to ensure high-quality feature extraction.\\
\noindent\textbf{Video-based Human Reconstruction.}\hspace{1em}
Video-based methods aim to balance per-frame accuracy with temporal consistency. Early works like HMMR \cite{art-15} and VIBE \cite{art-27} utilized Recurrent Neural Networks (GRUs) and adversarial training (motion discriminators) to impose kinematic priors. To capture longer-range dependencies, recent approaches have shifted towards Transformer-based architectures. MEVA \cite{art-43} and TCMR \cite{art-44} explore VAEs and temporal attention to refine motion sequences. State-of-the-art methods such as GLoT \cite{art-20}, and more recent works like TRAM \cite{tram-2024}, leverage global trajectory modeling and masked motion completion to achieve impressive smoothness and global consistency. However, these methods typically rely on implicit attention mechanisms to aggregate temporal cues. They assume that input features from the video sequence are relatively reliable. In extreme scenarios characterized by severe motion blur or heavy occlusion (\eg, diving, gymnastics), visual features often collapse. Under such conditions, implicit aggregation risks propagating noise from degraded frames rather than recovering them. In contrast, our approach adopts an explicit anchor-guided strategy, prioritizing the propagation of high-quality features to repair degraded segments.\\
\noindent\textbf{Action Quality Assessment (AQA).}\hspace{1em}
AQA aims to quantify the quality of action execution. While traditional methods rely on raw video appearance, recent state-of-the-art approaches (\eg, Zhu \etal \cite{art-21}) demonstrate that fusing appearance with explicit human pose sequences significantly boosts performance. We adopt this fusion-based AQA task as a downstream evaluation metric. This allows us to assess not just the geometric accuracy (MPJPE), but the temporal stability and perceptual quality of our reconstructed motion in complex, real-world environments.
\begin{figure}[!htb]
	\centering
	\includegraphics[width=1\linewidth]{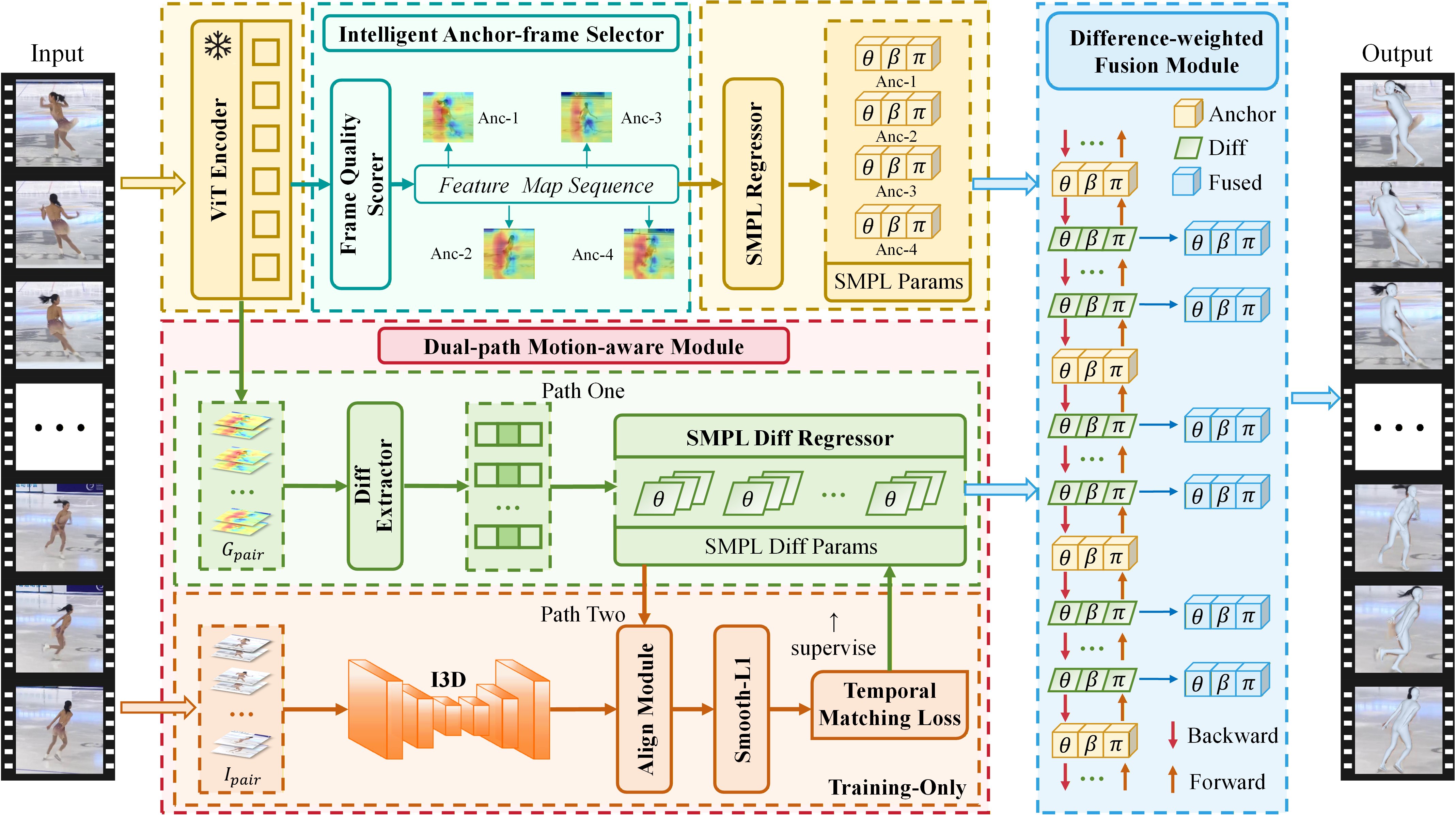}
	\caption{
		\textbf{Overview of the Again-Pose framework.} 
		Designed to handle extreme motion degradation, the pipeline consists of three stages: 
		(1) Given an input video sequence $I$, the \textbf{Intelligent Anchor-frame Selector} explicitly evaluates feature quality to identify reliable Anchor Frames, filtering out those suffering from feature collapse. 
		(2) The \textbf{Dual-path Motion-aware Module} captures inter-frame dynamics: \textit{Path One} models fine-grained parametric differences ($\Delta\theta$) for inference, while \textit{Path Two} (training-only) utilizes I3D features to enforce visual-motion consistency. 
		(3) The \textbf{Difference-weighted Fusion Module} propagates these reliable motion cues from anchors to "inpaint" the intermediate frames, adaptively fusing forward and backward predictions to reconstruct the complete, robust SMPL sequence.
	}
	\label{fig:Framework}
\end{figure}
\section{Method}
\subsection{Overview}
\noindent Our approach builds on the classic SMPL parameterized human model \cite{art-1}. The model takes a low-dimensional pose parameter ($\theta\in\mathbb{R}^{24\times3\times3}$) and a shape parameter ($\beta\in\mathbb{R}^{10}$) as input, and outputs a 3D mesh $M\in\mathbb{R}^{3\times N}$ with $N=6890$ vertices. The pose parameter $\theta$ comprises a global orientation $\theta_g\in\mathbb{R}^3$ and a local body pose $\theta_b\in\mathbb{R}^{23\times3}$. The final 3D joint locations $X\in\mathbb{R}^{3\times k}$ are obtained from the mesh vertices via a predefined linear regressor $W$, namely $X=MW$. Although the latest models such as SKEL \cite{art-2} incorporate biomechanical constraints, they lack a mature evaluation system. To leverage the mature SMPL ecosystem and datasets, we therefore adopt the SMPL model as our human representation.\\
\indent The overall pipeline is illustrated in \cref{fig:Framework}. Given an RGB video sequence $I={\{I_t\}}_{t=1}^T$ containing $T$ frames, we process it through two independent branches. In the first branch, we first extract the feature map sequence $G={\{G_t\}}_{t=1}^T$ from $I$ using a ViT. Then, we use an Intelligent Anchor-frame Selector to filter out representative high-quality anchor-features from $G$. Based on the selected features, we estimate the corresponding SMPL parameters for the Anchor Frames using an SMPL Regressor. In the second branch, the feature map sequence from the ViT and the original image sequence are reconstructed into frame pairs, denoted as $I_{pair}={\{\{I_{t-1},I_{t}\}\}}_{t=2}^T$ and $G_{pair}={\{\{G_{t-1},G_{t}\}\}}_{t=2}^T$ respectively. These are then input into a difference feature extractor and an I3D network to extract difference features and motion appearance features, respectively. Next, the regression of the SMPL difference parameters is mainly handled by the SMPL Diff Regressor, which is further refined by aligning and supervising with the motion appearance features. Finally, the outputs from the two branches are integrated through the Difference-weighted Fusion Module to generate the final SMPL parameter sequence, denoted as $O={\{\{\theta_t,\beta_t,\pi_t\}\}}_{t=1}^T$.
\subsection{Architecture}
\noindent We introduce the technical details of the three core components (Intelligent Anchor-frame Selector, Dual-path Motion-aware Module, Difference-weighted Fusion Module) of Again-Pose and briefly overview the parameter regression method for the Anchor Frame at the end.\\
\noindent\textbf{Intelligent Anchor-frame Selector.}\hspace{1em}
This module is designed to address the undesirable effects of sudden changes in low-quality frames on overall sequence reconstruction. It introduces Anchor frames—high-quality, representative frames whose features are more salient compared to other frames, are more likely to generate reliable and accurate SMPL parameters—and selects them via an automated strategy. The procedure consists of two steps:\\
\indent First, we compute a per-frame quality score $s_{frame}$. Let $G \in \mathbb{R}^{T \times N \times D}$ denote the feature map sequence extracted by the ViT backbone, where $N=192$ and $D=1024$. The quality score is composed of two measures: (1) a base saliency $s_{base}$, and (2) a dynamic regression weight $w_{dynamic}$. To capture the relative importance of each frame within the temporal context, we apply a Softmax function along the temporal dimension $T$ of $G$, and then calculate the L2 norm for each frame:
\begin{equation}
	s_{base}^t = \|\text{Softmax}_{temporal}({G_t})\|_2.
\end{equation}
\indent The dynamic weight is predicted by a lightweight MLP $g(\cdot)$, defined as $w_{dynamic}=\sigma(g(F_{smpl}))$, where $\sigma$ denotes the softmax function and $F_{smpl}$ is the token vector obtained from the Transformer Decoder in the SMPL Regressor. The final score is a weighted fusion:
\begin{equation}
	s_{frame} = \lambda \cdot s_{base} + (1-\lambda) \cdot w_{dynamic},
\end{equation}
where $\lambda$ is a balancing hyperparameter empirically set to 0.3.\\
\indent Second, the final Anchor Frame set is determined by a filter-and-fill algorithm governed by two hyperparameters, TOP-K and MIN-DISTANCE. We first take the TOP-K highest-scoring frames from the sequence $S={\{s_{frame}^t\}}_{t=1}^T$ as candidates. Then, we traverse these candidates in descending score order and discard any frame whose temporal distance to any already selected Anchor Frame is below the MIN-DISTANCE threshold, yielding an initial Anchor Frame set. This screening may create overly long blank intervals, which can amplify interpolation error and hinder optimization. To address this, we design a recursive filling mechanism (Algorithm implementation refers to \textbf{suppl. S-3}) that recursively detects excessively long blank intervals in the time series. It selects frames with the highest local scores within these intervals on the condition that the MIN-DISTANCE parameter is satisfied, thereby avoiding excessively long blank intervals to prevent them from impairing the model optimization while maintaining the quality of Anchor Frames. It should be noted that this paper transforms the problem of defining the specific length of excessively long blank intervals into the problem of determining the minimum distance between Anchor Frames. Specifically, once the optimal MIN-DISTANCE is determined, we must ensure that at least one key frame is selected within an interval of a maximum of twice the MIN-DISTANCE plus 2 (since Anchor Frames already exist at both ends of the interval). As shown in \cref{tab:param ablation}, the optimal value of MIN-DISTANCE is 3, so the length of excessively long blank regions is set to 8.\\
\noindent\textbf{Dual-path Motion-aware Module.}\hspace{1em}
This module captures inter-frame dynamics via two pathways to ensure smooth, consistent transitions.\\
\indent (1) \textit{Path One (Parameter-difference Path).} Focusing on inference, this path extracts fine-grained pose differences. A cross-attention-based Diff Extractor takes feature map pairs ($G_{pair}$) and regresses dynamic confidence weights to adaptively aggregate spatial features. The decoder then extracts difference vectors, mapped to SMPL increments ($\Delta\theta, \Delta\beta, \Delta\pi$) via an MLP, while a multi-head self-attention mechanism integrates global temporal context to suppress jitter.\\
\indent (2) \textit{Path Two (Visual Detail Supervision Path).} To enforce visual-kinematic consistency, this path leverages the I3D network \cite{art-7} pre-trained on Kinetics \cite{art-7} to extract appearance features ($F_{I3D}$) rich in action dynamics. These features are projected into a shared latent space with the SMPL difference sequence $D$ via MLPs. Subsequently, we employ cascaded cross-attention (to extract modal differences) and self-attention mechanisms to fuse them into aligned representations ($F_{motion-enc}, F_{visual-enc}$), which are supervised by the Temporal Matching Loss. Note that for inference efficiency, visual features are cached and reused, avoiding redundant computation.\\
\noindent\textbf{Difference-weighted Fusion Module.}\hspace{1em}
\begin{figure}[!htb]
	\vspace{5pt}
	\centering
	\includegraphics[width=0.6\linewidth]{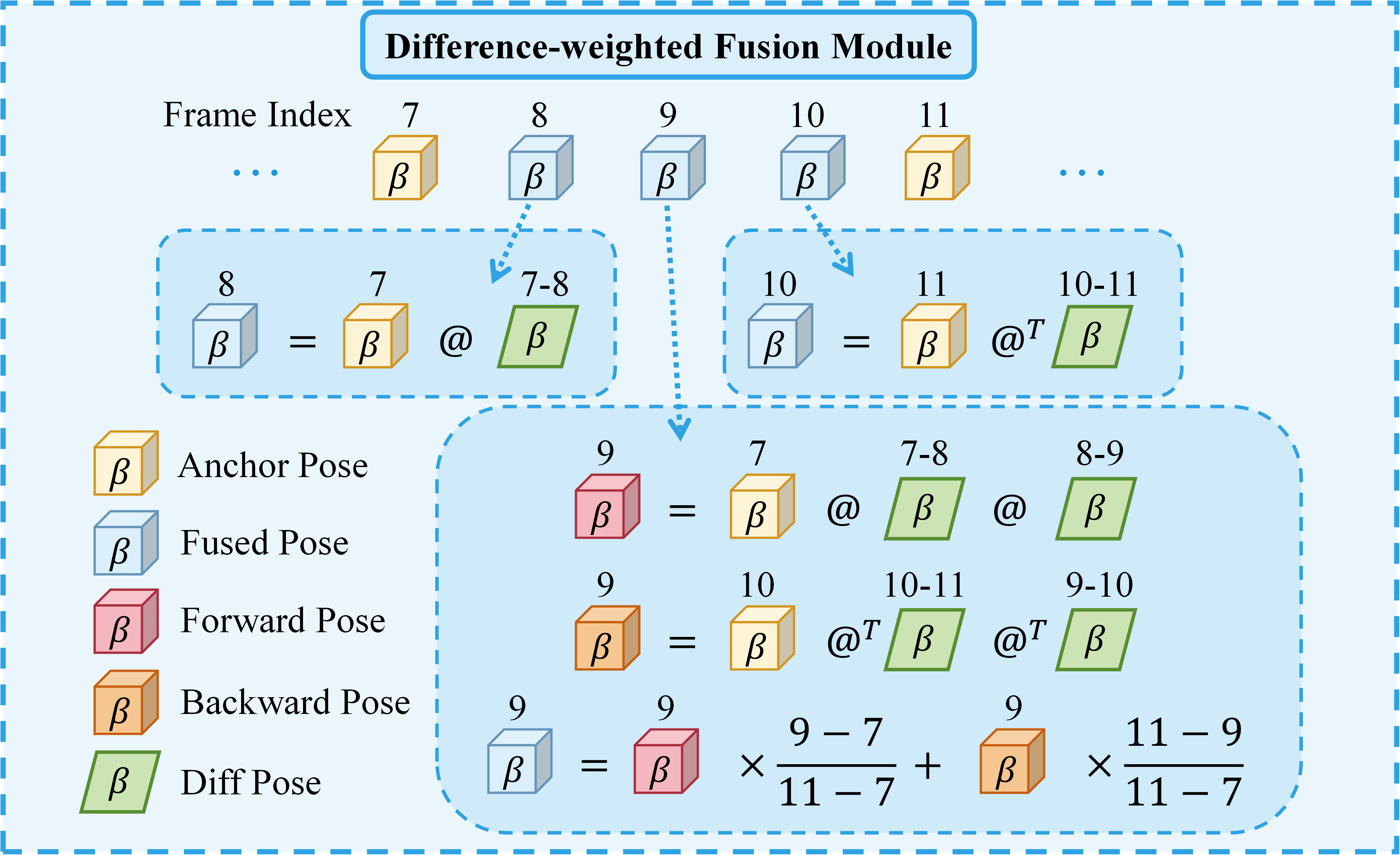}
	\caption{
		\textbf{Drift Suppression via Difference-weighted Fusion.} 
		To mitigate accumulation error, this module employs a bidirectional propagation strategy with a specific focus on transition areas.
		(1) \textbf{Bidirectional Propagation:} Motion differences are accumulated forward (\textcolor{red}{red}) and backward (\textcolor{orange}{orange}) from respective anchors via matrix multiplication (@). 
		(2) \textbf{Overlap-Specific Fusion:} We explicitly define an \textbf{Overlap Region} (controlled by OVERLAP-SIZE) between adjacent anchor trajectories. **Only frames within this region** undergo weighted fusion of the forward and backward paths; frames outside it rely on unidirectional propagation from the nearest anchor.
		(3) \textbf{Geometric Continuity:} Inside the overlap, to ensure stable linear interpolation, rotation matrices are transformed into the continuous \textbf{6D rotation representation} (Rot6D) before performing the distance-based weighted fusion.
	}
	\label{fig:Fusion}
\end{figure}
Unidirectional difference propagation may introduce systematic bias. To mitigate this issue, we adopt a difference-weighted fusion strategy in the overlap between Anchor Frame trajectories. This strategy uses a hyperparameter named OVERLAP-SIZE to control the size of the overlap. Since direct linear interpolation on rotation matrices or axis-angles is problematic, we convert the pose parameters $\theta$ into the continuous 6D rotation representation \cite{zhou2019continuity} before fusion, and convert them back to rotation matrices afterwards. For any frame inside an overlap, the final pose is a weighted combination of the two propagated estimates from the left and right Anchor Frames. The weights depend on the frame’s relative temporal distance to the two Anchor Frames. A frame closer to the left Anchor Frame receives a higher weight from the left path, and vice versa. This fusion reduces drift and ensures smooth transitions. \cref{fig:Fusion} illustrates the process.\\
\noindent\textbf{Anchor-frame Regression.}\hspace{1em}We directly regress SMPL parameters for all Anchor Frames with a dedicated SMPL Regressor. The regressor takes the Anchor Frames’ feature maps as input. It uses a Transformer-Decoder as the core. It leverages strong sequence modeling to decode pose and shape accurately.
\subsection{Training}
\noindent\textbf{Full-sequence Supervision Strategy.}\hspace{1em}
To ensure sufficient training of the single-frame parameter regressor (SMPL Regressor) and the difference parameter regressor (SMPL Diff Regressor), we adopt a full-sequence supervision strategy. Specifically, although during inference we only regress parameters directly from these Anchor Frames and interpolate between them, during training we utilize ground truth from the entire sequence for strong supervision: computing the single-frame prediction loss for all frames, as well as the difference prediction loss for all consecutive frame pairs. This approach ensures robustness of both regressors for their respective tasks, establishing a solid foundation for the combined performance during inference.\\
\noindent\textbf{Freezing Training.}\hspace{1em}
To enhance training efficiency and quickly validate the effectiveness of the core method, we freeze the backbone network (ViT) during training.

\subsection{Losses}
\noindent We adopt five loss functions to optimize the overall framework, three of which are derived from methods \cite{art-3, art-4, art-8} that demonstrate excellent performance. The five loss functions are as follows: 3D Joint Loss, 2D Joint Loss, SMPL Parameter Loss, Temporal Matching Loss, and Anchor-weight Loss. Next, we detail the calculation method of each loss.\\
\noindent\textbf{3D Joint Loss.}\hspace{1em}
We supervise predicted 3D joint locations with an L1 loss:
\begin{equation}
	L_{kp3D}=\left\|X_{true}^{3D}-X_{pred}\right\|,
\end{equation}
where $X_{true}^{3D}$ is the ground truth 3D joints and $X_{pred}$ is the prediction.\\
\noindent\textbf{2D Joint Loss.}\hspace{1em}
We supervise the 2D projections of the predicted 3D joints $\pi\left(X_{pred}\right)$ with an L1 loss:
\begin{equation}
	L_{kp2D}=\left\|X_{true}^{2D}-\pi\left(X_{pred}\right)\right\|.
\end{equation}
\noindent\textbf{SMPL Parameter Loss.}\hspace{1em}
We supervise pose $\theta$ and shape $\beta$ with MSE:
\begin{equation}
	L_{SMPL}={\left\|\theta_{true}-\theta_{pred}\right\|}^2+{\left\|\beta_{true}-\beta_{pred}\right\|}^2.
\end{equation}
\noindent\textbf{Temporal Matching Loss.}\hspace{1em}
We enforce consistency between motion and visual paths using the standard Smooth L1 Loss:
\begin{equation}
	L_{temp} = \frac{1}{N} \sum_{i} \Psi(x_i - y_i),
\end{equation}
where $x = F_{motion-enc}$, $y = F_{visual-enc}$, and $\Psi(d) = 0.5d^2$ if $|d| < 1$, otherwise $|d| - 0.5$. This robustness to outliers stabilizes the alignment.\\
\noindent\textbf{Anchor-weight Loss.}\hspace{1em}
To guide the model to assign higher dynamic weights to Anchor Frames with smaller expansion loss, we design the Anchor-weight Loss:
\begin{equation}
	L_{anchor} = \sum_{k \in \mathcal{A}} w_{dynamic}^{(k)} \cdot L_{anchor-extend}^{(k)},
\end{equation}
where $\mathcal{A}$ denotes the set of selected Anchor Frames, and $L_{anchor-extend}^{(k)}$ represents the total reconstruction loss for the subsequence covered by anchor $k$.
\section{Experiments}
\subsection{Datasets and Metrics}
\noindent\textbf{Datasets.}\hspace{1em}We use the training data provided by HMR2.0 \cite{art-3}. This training data is constructed by cropping and combining multiple public datasets, including Human3.6M \cite{art-9}, MPI-INF-3DHP \cite{art-10}, COCO \cite{art-11}, MPII \cite{art-12}, AI Challenger \cite{art-13}, AVA \cite{art-14}, and InstaVariety \cite{art-15}. For pose estimation benchmarks, we evaluate on three datasets with video sequences: Human3.6M \cite{art-9}, 3DPW \cite{art-16}, and PoseTrack \cite{art-17}. Among these, PoseTrack \cite{art-17} provides only 2D pose annotations. To evaluate the performance of Again-Pose in downstream tasks, we further conduct experiments on the FineDiving \cite{art-18} dataset for AQA tasks. In ablation studies, we use Human3.6M \cite{art-9} as the dataset.\\
\noindent\textbf{Metrics.}\hspace{1em}We adopt three common metrics for pose estimation: MPJPE (Mean Per Joint Position Error), PA-MPJPE (Procrustes-aligned MPJPE), and PCK (Percentage of Correct Keypoints). MPJPE and PA-MPJPE evaluate 3D pose accuracy, while PCK measures 2D pose accuracy. For AQA, we use $\rho$ (Spearman’s rank correlation coefficient), $R_{\ell2}$ (Relative $\ell2$-distance), and AIoU (Average Intersection Over Union) \cite{art-18}.
\begin{table}[tb]
	\centering
	\scriptsize
	\caption{Comparison with state-of-the-art single-frame-based approaches that regress SMPL parameters. *: trained on the 3DPW dataset.}
	\begin{tabular}{l|cc|cc|cc}
		\toprule
		\multirow{2}{*}{Methods} & \multicolumn{2}{c|}{PoseTrack} & \multicolumn{2}{c|}{3DPW} & \multicolumn{2}{c}{Human3.6M} \\
		\cmidrule{2-7}
		& @0.05$\uparrow$ & @0.1$\uparrow$ & MPJPE$\downarrow$ & PA-MPJPE$\downarrow$ & MPJPE$\downarrow$ & PA-MPJPE$\downarrow$ \\
		\midrule
		PARE~\cite{art-22} & 0.79 & 0.93 & 82.0 & 50.9 & 76.8 & 50.6 \\
		CLIFF~\cite{art-23} & 0.75 & 0.92 & --~* & --~* & \textbf{47.1} & 32.7 \\
		HyrbIK~\cite{art-24} & 0.81 & 0.94 & 80.0 & 48.8 & 54.4 & 34.5 \\
		PLIKS~\cite{art-25} & 0.74 & 0.94 & --~* & --~* & 47.0 & 34.5 \\
		HMR2.0~\cite{art-3} & \textbf{0.90} & \textbf{0.98} & 81.3 & 54.3 & 50.0 & 32.4 \\
		HSMR~\cite{art-2} & \textbf{0.90} & \textbf{0.98} & 81.5 & 54.8 & 50.4 & 32.9 \\
		\midrule
		\textbf{Ours} & \textbf{0.90} & \textbf{0.98} & \textbf{70.7} & \textbf{44.5} & 48.8 & \textbf{30.7} \\
		\bottomrule
	\end{tabular}
	\label{tab:comparison with frame-based}
\end{table}
\begin{table}
	\centering
	\scriptsize
	\caption{Comparison with state-of-the-art video-based approaches that regress SMPL parameters. *: trained on the 3DPW dataset.}
	\begin{tabular}{l|cc|cc}
		\toprule
		\multirow{2}{*}{Methods} & \multicolumn{2}{c|}{3DPW} & \multicolumn{2}{c}{Human3.6M} \\
		\cmidrule{2-5}
		& MPJPE$\downarrow$ & PA-MPJPE$\downarrow$ & MPJPE$\downarrow$ & PA-MPJPE$\downarrow$ \\
		\midrule
		DSD~\cite{art-26} & - & 69.5 & 59.1 & 42.4 \\
		VIBE~\cite{art-27} & 93.5 & 56.5 & 65.9 & 41.5 \\
		GLoT~\cite{art-20} & 80.7* & 50.6* & 67.0 & 46.3 \\
		\midrule
		\textbf{Ours} & \textbf{70.7} & \textbf{44.5} & \textbf{48.8} & \textbf{30.7} \\
		\bottomrule
	\end{tabular}
	\label{tab:comparison with video-based}
	\vspace{-3mm}
\end{table}
\subsection{Implement Details}
\noindent We use the same method as HMR 2.0 \cite{art-3} to load pre-trained weights into ViT. I3D loads the weights pre-trained on the Kinetics dataset \cite{art-7}. For optimization, we use the AdamW optimizer with a learning rate of 1e-5, a weight decay of 1e-4, and a batch size of 32. The key hyperparameters are set as follows: TOP-K is 6, MIN-DISTANCE is 3, OVERLAP-SIZE is 1, and the input sequence length is fixed at 16. All models are trained on a single H800 (80GB) GPU.
\begin{table}[t]
	\centering
	\scriptsize
	\caption{Comparison of the performance of several state-of-the-art human body reconstruction methods on AQA task.}
	\begin{tabular}{l|cc|c}
		\toprule
		Methods & $\rho\uparrow$ & $R_{\ell2}\downarrow(\times100)$ & AIoU@0.75$\uparrow$ \\
		\midrule
		Base~\cite{art-21} & 0.9465 & 0.2243 & 0.9841 \\
		Base+HMR 2.0~\cite{art-3} & 0.9421 & 0.2301 & 0.9655 \\
		Base+HSMR~\cite{art-2} & 0.9474 & 0.2199 & 0.9761 \\
		\midrule
		\textbf{Ours} & \textbf{0.9503} & \textbf{0.2113} & \textbf{0.9902} \\
		\bottomrule
	\end{tabular}
	\label{tab:comparison on AQA}
	\vspace{-3mm}
\end{table}
\subsection{Comparison with state-of-the-art methods}
\noindent\textbf{Single-frame-based methods.}\hspace{1em}We comprehensively compare Again-Pose with state-of-the-art methods on PoseTrack \cite{art-18}, 3DPW \cite{art-17}, and Human3.6M \cite{art-9}. \textbf{Protocol Setup:} Following the standard protocol for generalization evaluation \cite{art-3, art-10}, Again-Pose is \textit{not} trained on the 3DPW dataset. To ensure a fair comparison, we primarily compare against methods that also adhere to this protocol (\ie, excluding methods that fine-tune on 3DPW, such as CLIFF \cite{art-22} in some settings). As shown in \cref{tab:comparison with frame-based}, Again-Pose outperforms existing single-frame-based methods under this strict generalization setting. It is worth noting that while CLIFF \cite{art-22} achieves strong performance, it benefits from integrating bounding box location priors and, in many reported results, includes 3DPW in its training set. In contrast, our method achieves superior performance on the challenging 3DPW dataset (\textbf{70.7mm}) compared to other methods that do not see 3DPW during training (\eg, HMR 2.0 at 81.3mm), highlighting our strong cross-dataset generalization capability. This result strongly demonstrates that our method effectively leverages temporal information and inter-frame differences to compensate for deficiencies caused by low-quality frames, improving overall robustness without relying on domain-specific training data.\\
\noindent\textbf{Video-based methods.}\hspace{1em}Similarly, we compare Again-Pose with temporal methods on 3DPW and Human3.6M. As shown in \cref{tab:comparison with video-based}, Again-Pose outperforms established methods like VIBE \cite{art-27} and GLoT \cite{art-20}. We also acknowledge recent global trajectory methods like TRAM \cite{tram-2024}. While TRAM excels in world-space navigation via motion smoothing, our work targets a fundamentally different challenge: \textbf{local pose fidelity under extreme degradation} (\eg, severe blur in sports). Methods relying on implicit smoothing priors risk over-smoothing rapid acrobatic motions when visual features collapse. In contrast, Again-Pose explicitly propagates high-quality anchors to recover detailed kinematics, offering a solution orthogonal to trajectory-focused approaches.
\subsection{Comparison with methods for AQA}
\noindent To validate the adaptability and robustness of our framework in extreme scenarios, we conduct performance comparisons between Again-Pose and other methods on the FineDiving dataset \cite{art-18} using the pose-based AQA evaluation method proposed by Zhu \etal \cite{art-21}. It should be noted that the original method incorporates 2D pose information. In this experiment, we modify the component that applies pose information to a 3D form, while keeping the rest unchanged.\\
\indent The experimental results are shown in \cref{tab:comparison on AQA}. HMR 2.0 \cite{art-3} causes sudden changes in detection due to its sensitivity to blurred frames, resulting in a decline in its AQA score. Although the HSMR \cite{art-2} method maintains pose rationality to some extent through biomechanical constraints, it fails to completely avoid sudden changes. Notably, these sudden change issues become particularly prominent in AIoU, an indicator highly sensitive to temporal stability. In comparison, Again-Pose achieves the best performance among the three methods. This result demonstrates that Again-Pose effectively suppresses sudden changes in blurred frames during high-speed motion scenarios, thereby enhancing the stability of the pose sequence and the applicability of downstream tasks.\\
\begin{table}[t]
	\centering
	\scriptsize
	\caption{Ablation study on different modules and mechanisms in our framework on the Human3.6M dataset.}
	\begin{tabular}{c|cccc|cc}
		\toprule
		Methods & \makecell{Anchor-frame\\Sel. Strategy} & \makecell{Overlap Fusion\\Method} & \makecell{Visual Detail\\Supervision Path} & \makecell{Full-sequence\\Supervision} & MPJPE$\downarrow$ & PA-MPJPE$\downarrow$ \\
		\midrule
		A & Random & ours & $\surd$ & $\surd$ & 51.2 & 32.1 \\
		B & Uniformly & ours & $\surd$ & $\surd$ & 50.8 & 31.9 \\
		C & ours & Avg & $\surd$ & $\surd$ & 49.5 & 31.2 \\
		D & ours & $\times$ & $\surd$ & $\surd$ & 49.9 & 31.5 \\
		E & ours & ours & $\times$ & $\surd$ & 55.2 & 36.8 \\
		F & ours & ours & $\surd$ & $\times$ & 52.4 & 32.7 \\
		\midrule
		G & ours & ours & $\surd$ & $\surd$ & 48.8 & 30.7 \\
		\bottomrule
	\end{tabular}
	\label{tab:main ablation}
\end{table}
\begin{table}[t]
	\centering
	\scriptsize
	\caption{Ablation study on the two important hyperparameters in our framework.}
	\label{tab:param ablation}
	\begin{tabular}{c|c|cc}
		\toprule
		Param Name & Size & MPJPE$\downarrow$ & PA-MPJPE$\downarrow$ \\
		\midrule
		\multirow{5}{*}{TOP-K} & 1 & 57.1 & 38.4 \\
		& 4 & 50.9 & 32.0 \\
		& \textbf{6} & \textbf{48.8} & \textbf{30.7} \\
		& 8 & 49.7 & 31.6 \\
		& 16 & 49.5 & 31.3 \\
		\midrule
		\multirow{3}{*}{MIN-DISTANCE} & 1 & 50.2 & 32.1 \\
		& \textbf{3} & \textbf{48.8} & \textbf{30.7} \\
		& 6 & 58.6 & 39.4 \\
		\bottomrule
	\end{tabular}
	\vspace{-2mm}
\end{table}
\subsection{Ablation Study}
\noindent We conduct systematic ablation studies on the Human3.6M \cite{art-9} dataset to quantitatively evaluate the contributions of core components in our proposed framework. \cref{tab:main ablation} compares the model performance after replacing or removing each module individually, thereby verifying the effectiveness of each component. Additionally, \cref{tab:param ablation} analyzes the impacts of two critical hyperparameters—TOP-K and MIN-DISTANCE—on model performance. \\
\indent The following provides detailed explanations of the designs and results of the ablation methods in \cref{tab:main ablation}: \\
\indent (1) Methods A and B replace the Anchor-frame selection strategy with random and uniform selection, respectively. Both approaches exhibit significant performance degradation, demonstrating the effectiveness of the proposed Intelligent Anchor-frame Selector; \\
\indent (2) Methods C and D substitute the overlap fusion method with average fusion and hard switching. Compared to the proposed distance-weighted fusion method, both alternative approaches show performance reduction, validating the superiority of the designed Difference-weighted Fusion Module; \\
\indent (3) Method E removes the Visual Detail Supervision Path provided by the I3D network, resulting in substantial performance decline, highlighting the critical role of this pathway in the Dual-path Motion-aware Module; \\
\indent (4) Method F eliminates the full-sequence supervised training strategy, restricting the parameter regressor to partial sequence data, which similarly causes performance degradation and illustrates the importance of complete sequence information for model training. \\
\indent In terms of hyperparameter influence, the TOP-K section of \cref{tab:param ablation} demonstrates that the setting of TOP-K values requires a trade-off: a small value increases the probability of selecting low-quality Anchor Frames, thereby reducing performance, while an excessively large value forces the inclusion of frames with relatively lower quality, which also negatively impacts performance. The MIN-DISTANCE section in \cref{tab:param ablation} indicates that the setting of MIN-DISTANCE also needs to balance two factors: if the value is too small, it will lose its design significance for the differential interpolation method proposed in this paper; conversely, if set too large, it will lead to the accumulation of errors during interpolation, increasing the difficulty of model optimization.
\subsection{Efficiency Analysis}
\noindent While Again-Pose achieves state-of-the-art robustness, the dual-path mechanism introduces computational overhead compared to single-frame baselines. Specifically, while Path Two processes frame pairs, the feature reuse mechanism designed within the Dual-path Motion-aware Module effectively halves the redundant computation during practical inference. Given the target application of high-fidelity reconstruction for offline analysis (\eg, AQA, film production), we prioritize reconstruction quality and stability over real-time performance, aligning with recent trends in high-precision 4D reconstruction \cite{sun2026align}. For a detailed quantitative breakdown of the network parameters, FLOPs, and the effectiveness of this mechanism, please refer to \textbf{suppl. S-1} of the supplementary material.
\begin{figure}[!htb]
	\centering
	\includegraphics[width=1\linewidth]{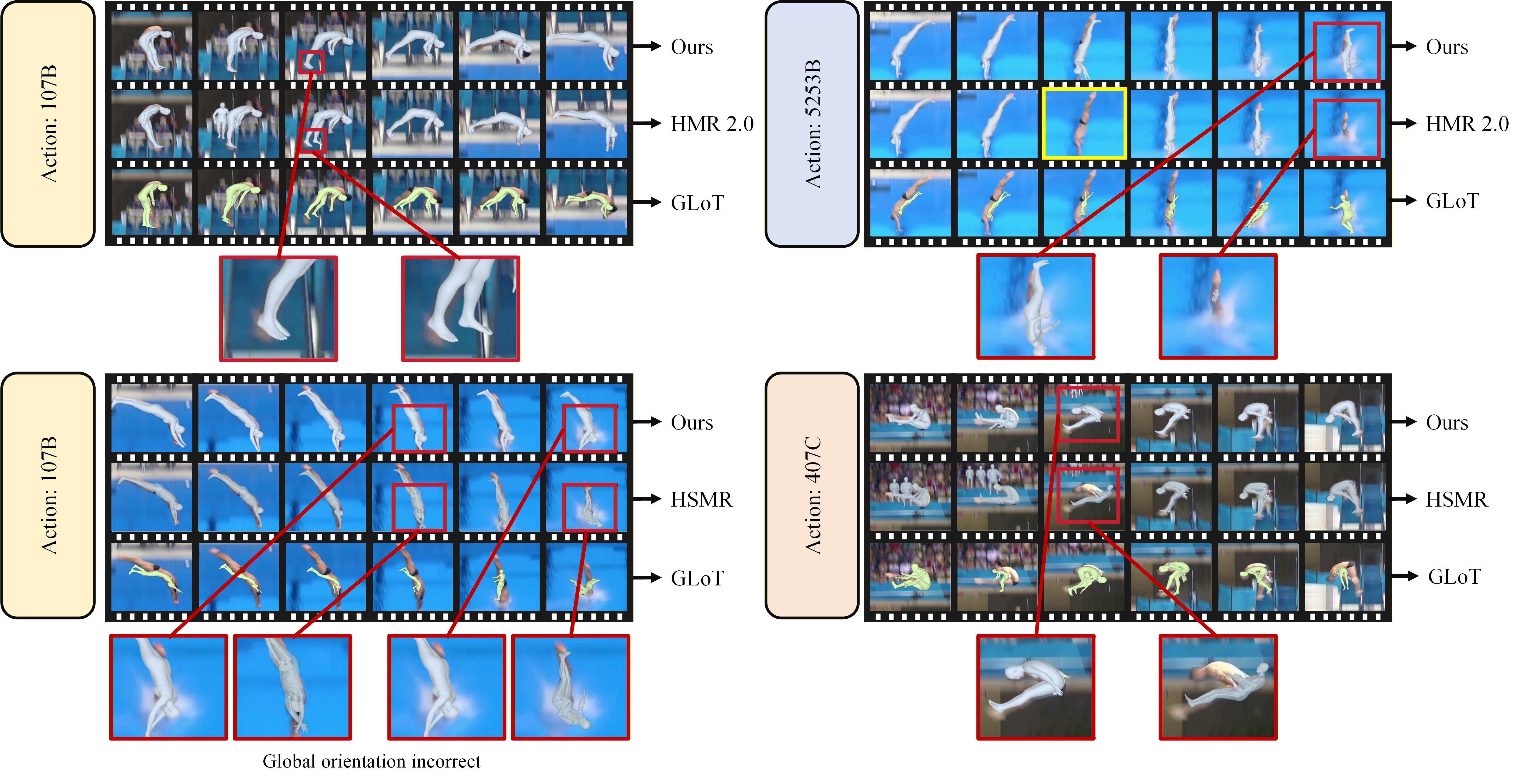}
	\caption{
		\textbf{Qualitative comparison on extreme diving actions (107B, 5253B, 407C).} 
		Visual results on the FineDiving dataset \cite{art-18} under severe motion blur and occlusion. We compare our Again-Pose framework with single-frame baselines (HMR 2.0 \cite{art-3} and HSMR \cite{art-2}) and a state-of-the-art video-based method (GLoT \cite{art-20}).
	}
	\label{fig:Vis}
\end{figure}
\subsection{Qualitative Analysis on Extreme Motion}
\noindent We visualize the reconstruction performance on the FineDiving dataset \cite{art-18} across diverse diving actions, including 107B, 5253B, and 407C. As shown in \cref{fig:Vis}, we compare Again-Pose against single-frame baselines (HMR 2.0 \cite{art-3}, HSMR \cite{art-2}) and the temporal method (GLoT \cite{art-20}).\\
\noindent\textbf{Comparison with GLoT.}\hspace{1em}It is evident that GLoT struggles significantly in this scenario. Due to the extreme motion blur, GLoT suffers from continuous feature collapse throughout the sequences, failing to produce plausible human meshes.\\
\noindent\textbf{Comparison with Single-frame Methods.}\hspace{1em}HMR 2.0 and HSMR perform reasonably well in stable phases. However, in moments of rapid rotation or self-occlusion, they frequently fail to detect limbs or suffer from severe jitter, producing unstable predictions.\\
\noindent\textbf{Stability in the Limit.}\hspace{1em}It is worth noting that reconstructing 4D humans in such extreme sports scenarios remains a highly challenging task. While our reconstruction results are not yet flawless in the most extreme frames, Again-Pose demonstrates superior stability compared to competing methods. By explicitly propagating high-quality anchors, our method effectively "inpaints" the missing details. As observed in the zoomed-in regions, Again-Pose successfully avoids the sudden structural collapse and severe jitter seen in HMR 2.0 and HSMR, maintaining structural integrity and providing a much more consistent and robust motion sequence (extended occlusion results are in \textbf{suppl. S-2}).
\section{Conclusion}
\noindent In this paper, we introduce Again-Pose, a robust framework designed to address the fragility of existing video-based pose estimation methods in extreme motion scenarios. We identify a critical failure mode in state-of-the-art temporal models: their reliance on implicit feature aggregation leads to model collapse when input frames suffer from severe degradation such as motion blur or occlusion. To overcome this, we propose an explicit anchor-guided strategy that decouples the video sequence into reliable anchors and reconstructible intervals. By leveraging a Dual-path Motion-aware Module, Again-Pose effectively propagates high-fidelity kinematic and visual cues to "inpaint" the poses of degraded frames, rather than blindly aggregating noisy features. Extensive experiments demonstrate that our approach not only achieves state-of-the-art performance on standard benchmarks (Human3.6M \cite{art-9}, 3DPW \cite{art-16}, PoseTrack \cite{art-17}) under strict cross-dataset generalization protocols but also exhibits superior robustness in the challenging FineDiving dataset \cite{art-18}. Furthermore, the improvements in the downstream AQA task validate that our reconstructed motions possess higher temporal stability and perceptual quality.

\clearpage
\appendix
\section*{Supplementary Material}
\setcounter{table}{0}
\renewcommand{\thetable}{S-\arabic{table}}
\renewcommand{\theHtable}{supp.\arabic{table}}
\setcounter{figure}{0}
\renewcommand{\thefigure}{S-\arabic{figure}}
\renewcommand{\theHfigure}{supp.\arabic{figure}}
\setcounter{section}{0}
\renewcommand{\thesection}{S-\arabic{section}}
\renewcommand{\theHsection}{supp.\arabic{section}}
\section{Computational Complexity and Efficiency Analysis}
\noindent To address the concerns regarding computational redundancy and runtime efficiency, we provide a detailed analysis of the computational complexity of Again-Pose compared to the state-of-the-art single-frame baseline (\eg, HMR 2.0 \cite{art-3}). As discussed in Section 4.6 of the main paper, a naive implementation of our Dual-path Motion-aware Module indeed leads to redundant feature extraction due to the overlapping frame pairs in Path Two. However, by implementing a strict Feature Caching Strategy, the feature map $G_t$ extracted by the heavy ViT backbone is cached and reused. This ensures that the backbone strictly processes each frame only once .The quantitative comparison is summarized in Table \cref{tab:complexity}.\\
\begin{table}[h]
	\vspace{-8mm}
	\centering
	\caption{Computational Complexity Comparison. $T$ denotes the sequence length.}
	\label{tab:complexity}
	\begin{tabular}{llccc}
		\toprule
		Method & Backbone & \makecell{Parameters (M)} & \makecell{FLOPs (G)\\per seq.} & \makecell{FLOPs (G)\\per frame (avg)} \\
		\midrule
		HMR 2.0 \cite{art-3} & ViT-H & 680 & $123 \times T$ & 123.0 \\
		\textbf{Again-Pose (Ours)} & ViT-H & \textbf{750} & \textbf{$123.6 \times T$} & \textbf{123.6} \\
		\bottomrule
	\end{tabular}
	\vspace{-12mm}
\end{table}
\subsection{Analysis of the Overhead}
\noindent\textbf{Marginal FLOPs Increase:}\hspace{1em}With the caching strategy, the computational bottleneck remains the ViT backbone. Our proposed temporal modules (Intelligent Anchor-frame Selector, Dual-path Motion-aware Module, and Difference-weighted Fusion Module) are highly lightweight. They add only 0.6G FLOPs per frame, representing a marginal ~0.48\% increase in computational cost compared to the single-frame baseline.\\
\noindent\textbf{Parameter Efficiency:}\hspace{1em}The total parameter count increases from 680M to 750M (+70M). This modest ~10\% increase is entirely attributed to the newly introduced cross-attention and self-attention layers designed for inter-frame motion extraction and visual detail supervision.\\
\noindent\textbf{Performance vs. Cost Trade-off:}\hspace{1em}Given the catastrophic failure of single-frame methods in extreme motion scenarios (as shown in the main paper's experiments), this negligible computational overhead yields a disproportionately large gain in structural stability and robustness. For downstream tasks like Action Quality Assessment (AQA) or offline motion analysis where high-fidelity reconstruction is paramount, this trade-off is highly favorable.
\section{Extended Qualitative Results on Extreme Sports}
\begin{figure}[!htb]
	\vspace{-3mm}
	\centering
	\includegraphics[width=1\linewidth]{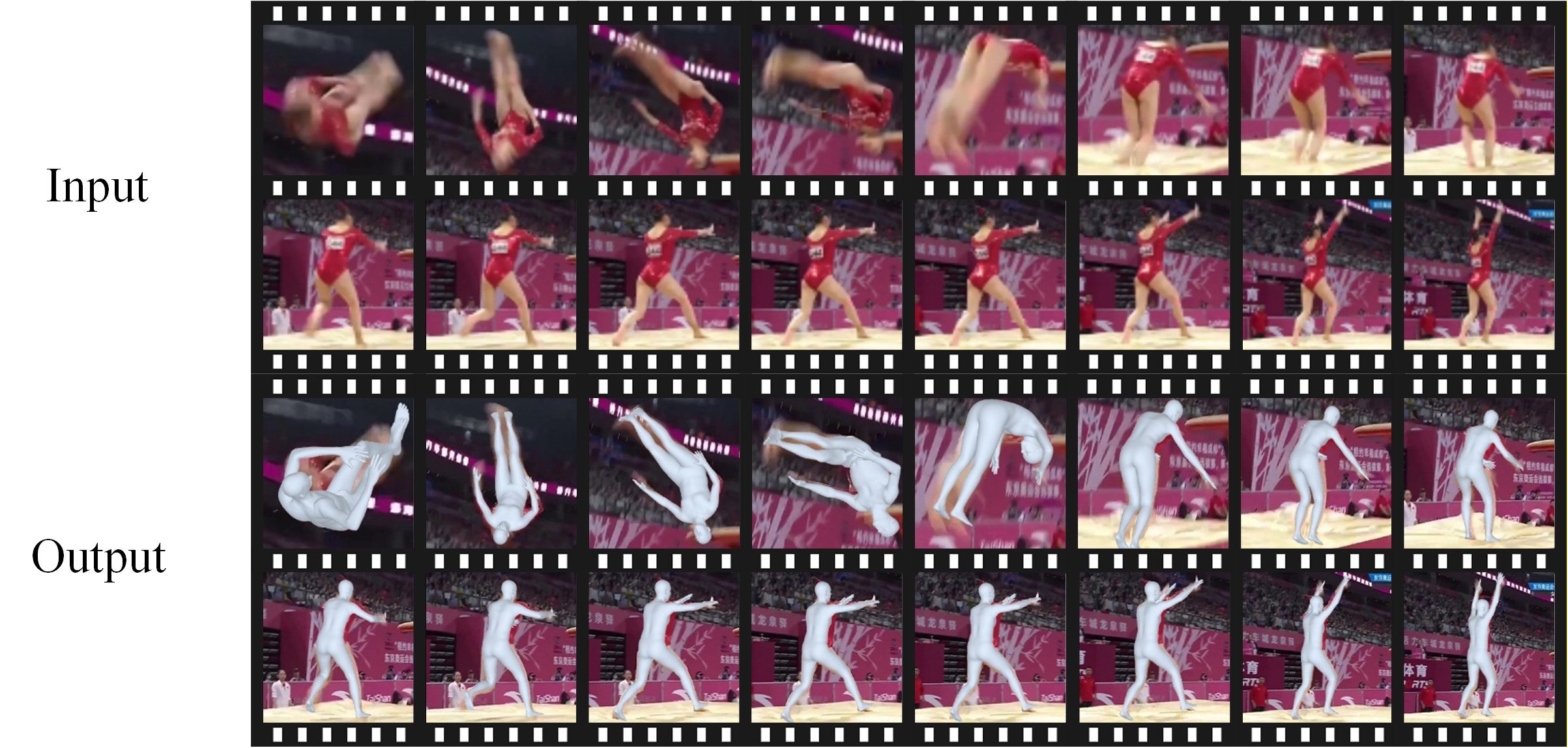}
	\caption{\textbf{Qualitative results on extreme gymnastics sequences.} Again-Pose successfully recovers biomechanically plausible 3D meshes under severe self-occlusions and rapid mid-air rotations.}
	\label{fig:vis-1}
\end{figure}

\begin{figure}[!htb]
	\centering
	\includegraphics[width=1\linewidth]{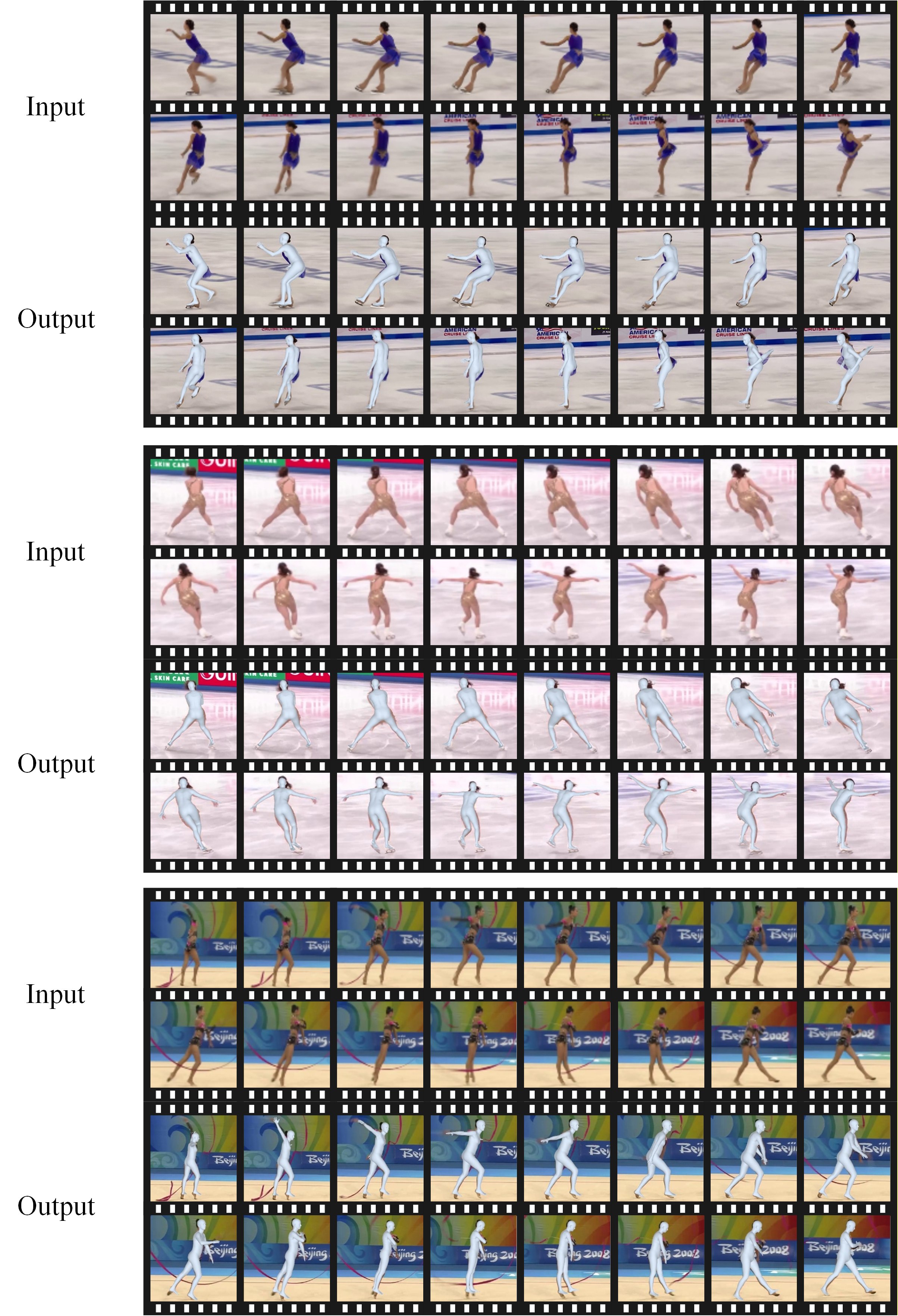}
	\caption{\textbf{High-fidelity reconstruction in mixed extreme sports.} Our method maintains structural integrity and smooth continuous tracking despite catastrophic motion blur.}
	\label{fig:vis-2}
\end{figure}

\begin{figure}[!htb]
	\centering
	\includegraphics[width=1\linewidth]{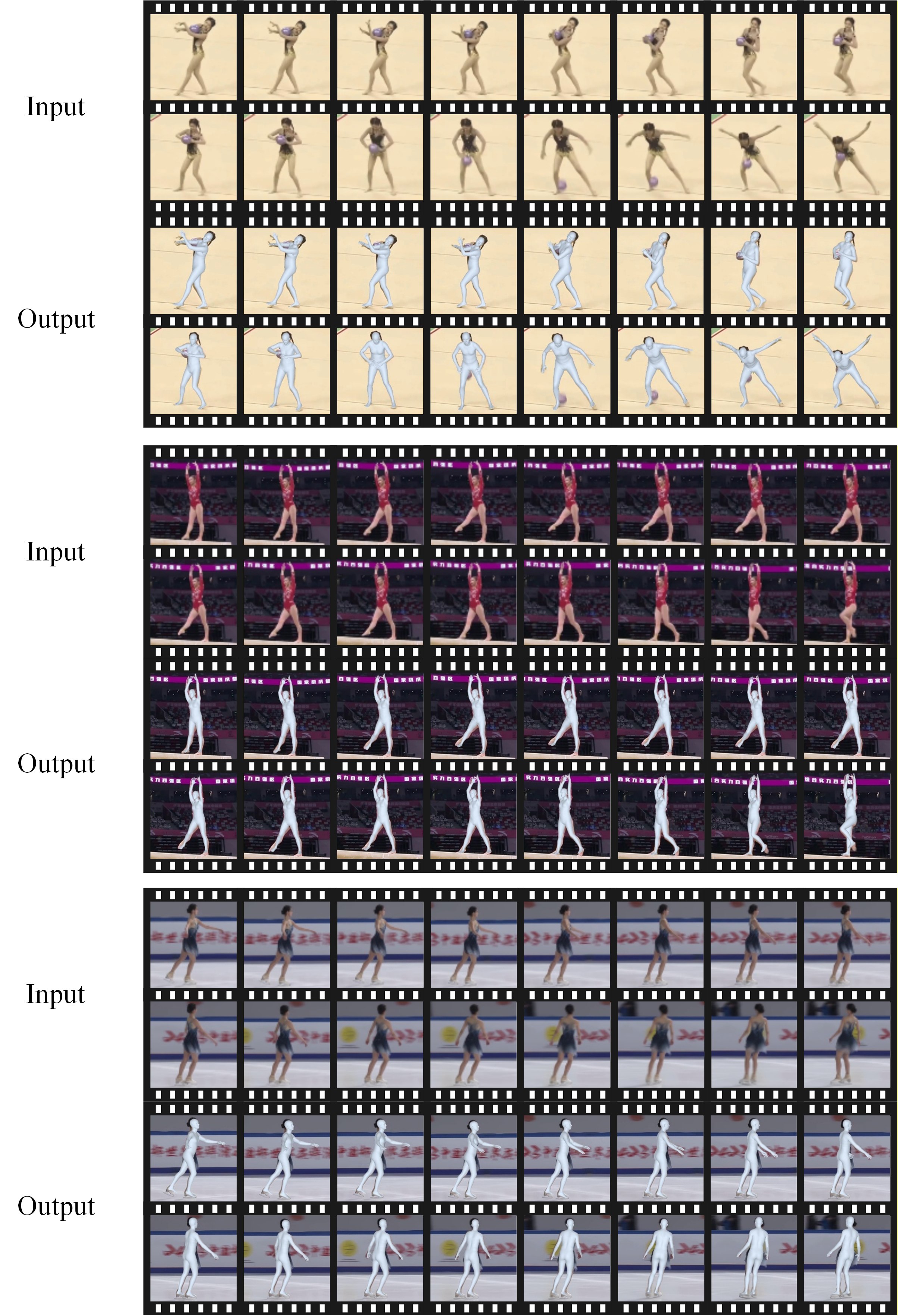}
	\caption{\textbf{Continuous tracking stability in unconstrained environments.} Again-Pose effectively suppresses accumulation drift, demonstrating strong generalization across highly diverse and complex acrobatic maneuvers.}
	\label{fig:vis-3}
\end{figure}

\noindent While Section 4.7 of the main paper provides rigorous head-to-head visual comparisons with state-of-the-art baselines, this section aims to demonstrate the broad generalization and continuous tracking stability of Again-Pose across a wider variety of unconstrained, high-speed sporting events.\\
\indent Specifically, we provide extended qualitative results on highly challenging sequences involving extreme gymnastics and figure skating. These scenarios are notorious for introducing severe visual degradation, including rapid full-body rotations, complex self-occlusions (\eg, tight mid-air tucks), and extreme motion blur that completely obscures local anatomical joints.\\
\indent As illustrated in Figures \cref{fig:vis-1}, \cref{fig:vis-2}, and \cref{fig:vis-3}, even when the input RGB frames suffer from catastrophic motion blur where the human silhouette is barely distinguishable, Again-Pose successfully reconstructs robust and biomechanically plausible 3D meshes. This stability is directly attributed to our explicit anchor-guided propagation mechanism and Difference-weighted Fusion Module. By explicitly relying on inter-frame dynamic cues anchored by adjacent high-quality frames rather than trusting degraded local visual features, our framework effectively "inpaints" the missing poses. These visualizations confirm that Again-Pose prevents error accumulation (drift) and maintains remarkable temporal smoothness and structural integrity throughout complex acrobatic maneuvers, addressing the common failure modes of existing temporal methods.
\section{Recursive Filling Algorithm}
\noindent To provide full implementation details for the "filter-and-fill" strategy described in Section 3.2 of the main paper, we present the core logic of our Recursive Filling Algorithm. This algorithm effectively mitigates excessively long blank intervals between Anchor Frames while strictly satisfying the \texttt{MIN-DISTANCE} constraint.
\begin{lstlisting}[
	style=academicsnippet,
	caption={Python implementation of the Recursive Filling Algorithm for Anchor Frame selection.}, 
	label={code:appendix-anchor}
	]
	anchor_indices = []
	for b_idx in range(batch_size):
	selected = sorted(filtered_anchors[b_idx])
	
	# Fill empty region in sequence header
	front_fill_empty(frame_quality_scores[b_idx], selected, selected[0])
	selected = sorted(selected)
	
	# Fill empty region in sequence tail
	behind_fill_empty(frame_quality_scores[b_idx], selected, selected[-1])
	selected = sorted(selected)
	
	# Fill empty regions in sequence middle
	new_selected = selected.copy()
	for i in range(len(selected) - 1):
	mid_fill_empty(frame_quality_scores[b_idx], new_selected, 
	selected[i], selected[i + 1])
	selected = sorted(new_selected)
	anchor_indices.append(selected)
	
	# ---------------------------------------------------------
	# Helper Functions for Recursive Filling
	# ---------------------------------------------------------
	def front_fill_empty(frame_quality_scores, selected, first_idx):
	"""Recursively fill sequence header"""
	empty_len = first_idx - min_distance + 1
	if empty_len > (min_distance // 2):
	block_indices = arange(0, empty_len)
	block_scores = frame_quality_scores[block_indices]
	best_idx = block_indices[argmax(block_scores)]
	selected.append(best_idx)
	front_fill_empty(frame_quality_scores, selected, best_idx)
	
	def behind_fill_empty(frame_quality_scores, selected, last_idx):
	"""Recursively fill sequence tail"""
	empty_len = sequence_length - (last_idx + min_distance)
	if empty_len > (min_distance // 2):
	block_indices = arange(sequence_length - empty_len, sequence_length)
	block_scores = frame_quality_scores[block_indices]
	best_idx = block_indices[argmax(block_scores)]
	selected.append(best_idx)
	behind_fill_empty(frame_quality_scores, selected, best_idx)
	
	def mid_fill_empty(frame_quality_scores, selected, prev_idx, next_idx):
	"""Recursively fill sequence middle"""
	empty_len = next_idx - prev_idx - 1
	if empty_len >= (min_distance * 2):
	begin_idx = prev_idx + min_distance
	end_idx = next_idx - min_distance
	block_indices = arange(begin_idx, end_idx + 1)
	block_scores = frame_quality_scores[block_indices]
	best_idx = block_indices[argmax(block_scores)]
	selected.append(best_idx)
	# Recurse on new sub-regions
	mid_fill_empty(frame_quality_scores, selected, prev_idx, best_idx)
	mid_fill_empty(frame_quality_scores, selected, best_idx, next_idx)
\end{lstlisting}
{
	\bibliographystyle{splncs04}
	\bibliography{main}
}
\end{document}